\documentclass[journal]{IEEEtran}

\usepackage{xspace}
\newcommand{\GoogleStaticMapsAPI}{Google Static Maps API\xspace}
\newcommand{\GoogleDirectionsAPI}{Google Maps Directions API\xspace}
\newcommand{\OpenStreetMap}{OpenStreetMap\xspace}
\newcommand{\IEEECopyrightMessage}{\copyright~2017 IEEE. Personal use of this material is permitted. Permission from IEEE must be obtained for all other uses, in any current or future media, including reprinting/republishing this material for advertising or promotional purposes, creating new collective works, for resale or redistribution to servers or lists, or reuse of any copyrighted component of this work in other works.\xspace}

\usepackage[noadjust]{cite}
\usepackage{hyperref}
\usepackage[usenames,dvipsnames]{xcolor}
\usepackage{amsmath}
\usepackage{tabularx, booktabs}
\usepackage{multirow}
\usepackage{bm}
\usepackage[utf8]{inputenc}
\usepackage[pdftex]{graphicx}

\definecolor{blue}{RGB}{41,5,195}
\makeatletter
\hypersetup{colorlinks=true,linkcolor=blue,citecolor=blue,urlcolor=blue}
\makeatother

\begin{document}
\pagenumbering{gobble}
	
\title{Deep Learning Based Large-Scale Automatic\\Satellite Crosswalk Classification}

\author{
	Rodrigo~F.~Berriel,
	André~Teixeira~Lopes,
	Alberto~F.~de Souza,
	and Thiago~Oliveira-Santos
}

\markboth{}
{Shell \MakeLowercase{\textit{et al.}}: Bare Demo of IEEEtran.cls for IEEE Journals}

\maketitle

\makeatletter
\def\ps@IEEEtitlepagestyle{
	\def\@oddfoot{\mycopyrightnotice}
	\def\@evenfoot{}
}
\def\mycopyrightnotice{
	{\footnotesize
		\begin{minipage}{\textwidth}
			\centering
			\scriptsize
			\IEEECopyrightMessage
		\end{minipage}
	}
}
\newcommand\copyrightnotice{%
	\begin{tikzpicture}[remember picture,overlay]
	\node[anchor=south,yshift=10pt] at (current page.south) {\fbox{\parbox{\dimexpr\textwidth-\fboxsep-\fboxrule\relax}{\copyrighttext}}};
	\end{tikzpicture}%
}

\begin{abstract}
	High-resolution satellite imagery have been increasingly used on remote sensing classification problems. One of the main factors is the availability of this kind of data. Even though, very little effort has been placed on the zebra crossing classification problem. In this letter, crowdsourcing systems are exploited in order to enable the automatic acquisition and annotation of a large-scale satellite imagery database for crosswalks related tasks. Then, this dataset is used to train deep-learning-based models in order to accurately classify satellite images that contains or not zebra crossings. A novel dataset with more than 240,000 images from 3 continents, 9 countries and more than 20 cities was used in the experiments. Experimental results showed that freely available crowdsourcing data can be used to accurately (97.11\%) train robust models to perform crosswalk classification on a global scale.
\end{abstract}


\IEEEpeerreviewmaketitle

\section{Introduction}

\begin{figure*}
	\centering
	\includegraphics[width=\textwidth]{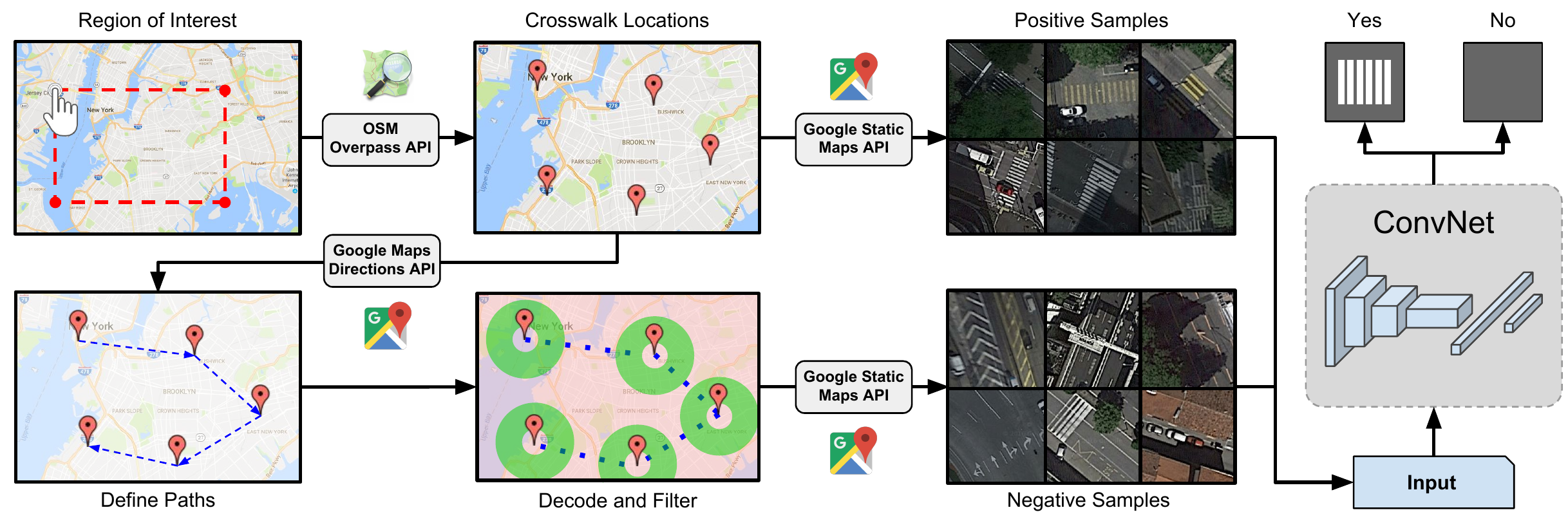}
	\caption{System architecture. The input is a region (red dashed rectangle) or a set of regions of interest. Firstly, known crosswalk locations (red markers) are retrieved using the OpenStreetMap (OSM). Secondly, using \GoogleDirectionsAPI, paths (blue dashed arrows) between the crosswalk locations are defined. Thirdly, these paths are decoded and the result locations are filtered (only locations within the green area are accepted) in order to decrease the amount of wrongly annotated images. At this point, positive and negative samples can be downloaded from \GoogleStaticMapsAPI. Finally, this large-scale satellite imagery is used to train Convolutional Neural Networks (ConvNets) to perform zebra crossing classification.}
	\label{fig:proposed-model}
\end{figure*}

\IEEEPARstart{Z}{ebra} crossing classification and detection are important tasks for mobility autonomy. Even though, there are few data available on where crosswalks are in the world. The automatic annotation of crosswalks' locations worldwide can be very useful for online maps, GPS applications and many others. In addition, the availability of zebra crossing locations on these applications can be of great use to people with disabilities, to road management, and to autonomous vehicles. However, automatically annotating this kind of data is a challenging task. They are often aging (painting fading away), occluded by vehicle and pedestrians, darkened by strong shadows, and many other factors.

A common approach to tackle these problems is using the camera on the phones to help the visually impaired people. Ivanchenko et al. \cite{ivanchenko2008detecting} developed a prototype of a cell phone application that determines if any crosswalk is visible and aligns the user to it. Their system requires some thresholds to be tuned, which may decrease performance delivered off-the-shelf. Another approach is a camera mounted on a car, usually aiming driver assistance systems. Haselhoff and Kummert \cite{haselhoff2010crosswalkadas} presented a strategy based on the detection of line segments. As discussed by the authors, their method is constrained to crosswalks perpendicular to the driving direction. Both perspectives (from a person \cite{ivanchenko2008detecting} or a car \cite{haselhoff2010crosswalkadas}) present a known limitation: there is a maximum distance in which crosswalks can be detected. Moreover, these images are quite different from those used in the work hereby proposed, i.e., satellite imagery. Nonetheless, Ahmetovic et al. \cite{ahmetovic2015zebra} presented a method combining both perspectives. Their algorithm searches for zebra crossings in satellite images. Subsequently, the candidates are validated against Google Street View images. One of the limitations is that the precision of the satellite detection step alone, as the one proposed in this work, is $\approx 20\%$ lower than the combined algorithm. In addition, their system takes 180ms to process a single image. Banich \cite{banich2016crosswalk} presented a neural-network-based model, in which the author used 11 features of stripelets, i.e., pair of lines. The author states that the proposed model cannot deal with crosswalks affected by shadows and can only detect white crosswalks. The aforementioned methods \cite{ivanchenko2008detecting, haselhoff2010crosswalkadas, ahmetovic2015zebra, banich2016crosswalk} were evaluated on manually annotated datasets that were usually local (i.e., within a single city or a country) and small (from 30 to less than 700 crosswalks).

Another approach that has been increasingly investigated is the use of aerial imagery, specially satellite imagery. Herumurti et al. \cite{herumurti2013urban} employed a circle mask template matching and SURF method to detect zebra crossing on aerial images. Their method was validated on a single region in Japan and the method that detects most crosswalks took 739.2 seconds to detect 306 crosswalks. Ghilardi et al. \cite{ghilardi2016crosswalk} presented a model to classify crosswalks in order to help the visually impaired. Their model is based on an SVM classifier fed with manually annotated crosswalk regions. As most of the other related works, the dataset used in their work is local and small (900 small patches, and 370 of crosswalks). Koester et al. \cite{koester2016zebra} proposed an SVM based on HOG and LBPH features to detect zebra crossings in aerial imagery. Although very interesting, their dataset (not publicly available) was gathered manually from satellite photos, therefore it is relatively small (3119 zebra crossings and $\approx12500$ negative samples) and local. In addition, their method shows a low generalization capability, because a model trained in one region shows low recall when evaluated in another (known as cross-based protocol), e.g., the recall goes from 95.7\% to 38.4\%.

In this letter, we present a system able to automatically acquire and annotate zebra crossings satellite imagery, and train deep-learning-based models for crosswalk classification in large scale. The proposed system can be used to automatically annotate crosswalks worldwide, helping systems used by the visually impaired, autonomous driving technologies, and others. This system is the result of a comprehensive study. This study assesses the quality of the available data, the most suitable model and its performance in several imagery levels (city, country, continent and global). In fact, this study is performed on real-world satellite imagery (almost 250,000 images) acquired and annotated automatically. Given the noisy nature of the data, results are also compared with human annotated data. The proposed system is able to train models that achieve 97.11\% on a global scale.

\section{Proposed Method}

The system comprises two parts: Automatic Data Acquisition and Annotation, and Model Training and Classification. An overview of the proposed method can be seen in the \autoref{fig:proposed-model}. Firstly, the user defines the regions of interest (regions where he wants to download crosswalks). The region of interest is given by the lower-left and the upper-right corners. After that, crosswalk locations within these regions are retrieved from the \OpenStreetMap\footnote{\url{http://www.openstreetmap.org}}. Subsequently, using the zebra crossing locations, positive and negative images (i.e., images that contain and do not contain crosswalks on it, respectively) are downloaded using the \GoogleStaticMapsAPI\footnote{\url{http://developers.google.com/maps/documentation/static-maps/}}. As the location of the crosswalks are known, the images are automatically annotated. Finally, these automatically acquired and annotate images are used to train a Convolutional Neural Network from the scratch to perform classification. Each process is described in details in the following subsections.

\subsection{Automatic Data Acquisition and Annotation}
To automatically create a model for zebra crossing classification, the first step is the image acquisition. Initially, regions of interest are defined by the user. These regions can either be defined manually or automatically (e.g. based on a given address, city, etc.). The region is rectangular (with the True North up) and is defined by four coordinate points: minimum latitude, minimum longitude, maximum latitude, maximum longitude (or South-West-North-East), i.e., the bottom-left and top-right corners (e.g. 40.764498, -73.981447, 40.799976, -73.949402 defines the Central Park, NY, USA region). For each region of interest, zebra crossing locations are retrieved from the OpenStreetMap (OSM) using the Overpass API\footnote{\url{http://wiki.openstreetmap.org/wiki/Overpass_API}}. Regions larger than 1/4 degree in either dimension are likely to be split into multiple regions, as OpenStreetMap servers may reject these requests. Even though, most of the settled part of the cities around the world meets this limitation (e.g. the whole city of Niterói, RJ, Brazil fits into a single region).

To download the zebra crossings, the proposed system uses the tag \texttt{highway=crossing} of the OSM, which is one of the most reliable and most used tag for this purpose. In possession of these crosswalk locations, the system needs to automatically find locations without zebra crossing to serve as negative samples. As simple as it may look, negative samples are tricky to find because of several factors. One of these is the relatively low coverage of the zebra crossing around the world. Another is the fact that crosswalks are susceptible to changes over the years. They may completely fade away, they can be removed and streets can change. Alongside these potential problems, OSM is built by volunteers, therefore open to contributions that may not be as accurate as expected. Altogether these factors indicate how noisy the zebra crossing locations may be. Therefore, picking up no-crosswalk locations must be done cautiously. The first step to acquire good locations for the negative samples is to filter only regions that contain roads. For that, the system queries \GoogleDirectionsAPI for directions from a crosswalk to another to ensure that the points will be within a road. In order to lower the number of requests to the \GoogleDirectionsAPI, our system adds 20 other crosswalks as waypoints between two crosswalk points (the API has a limit of up to 23 waypoints and 20 ensures this limit). The number of waypoints does not affect the accuracy performance of the final system since the same images would be downloaded but requiring more time. \GoogleDirectionsAPI responds to the request with an encoded polyline. The decoded polyline comprises a set of points in the requested path. The second step consists of virtually augmenting the number of locations using a fixed spacing of $1.5 \times 10^{-4}$ degrees (approximately 16 meters). All duplicate points are removed on the third step. Finally, the system also filters out all images too close or too far away. Too close locations may contain crosswalks and could create false positives; and too far locations may have non-annotated crosswalks and could create false negatives. Therefore, locations closer than $3 \times 10^{-4}$ degrees or farther than $6 \times 10^{-4}$ degrees or locations outside the region requested are removed to decrease the occurrence of false positives and false negatives.

After acquiring both positive and negative sample locations, the proposed system dispatches several threads to download the images using the \GoogleStaticMapsAPI. Each requested image is centered on the location to be requested. Also, the images are requested with a zoom factor equal to 20 and size of $200\times225$ pixels. This size was empirically defined to have a good trade-off between the image size and the field of view of the area. Bigger sizes would increase the probability of having crosswalks away of the queried location. As a result, each image covers $\approx 22 \times 25$ meters. Some positive and negative samples can be seen in the \autoref{fig:samples}.

\subsection{Model training and classification}

Before initializing the model training, an automatic pre-processing operation is required. Every image downloaded from the \GoogleStaticMapsAPI contains the Google logo and a copyright message on the bottom. In order to remove these features, 25 pixels are removed from the bottom of each image, cropping the original images from $200\times225$ to $200\times200$. In possession of all cropped images, both positive and negative samples, the training of the model can begin.

To tackle this large-scale problem, the proposed system uses a deep-learning-based model: a Convolutional Neural Network \cite{le1990handwritten}. 
The architecture of the model used by the system is the VGG \cite{simonyan2014very}. It was chosen after the evaluation of three different architectures: AlexNet \cite{krizhevsky2012imagenet}, VGG and GoogLeNet \cite{szegedy2015going} with 5, 16, and 22 convolutional layers, respectively. All models started from pre-trained models on the ImageNet \cite{ILSVRC15}, i.e., fine-tuning. In addition, the input images were upsampled from $200\times200$ to $256\times256$ using bilinear interpolation and the subtraction of the mean of the training set was performed. More details on the training configuration are described in the \autoref{sec:experiments}. Also, the last layer of VGG was replaced by a fully-connected layer comprising two neurons with randomly initialized weights, one for each class (crosswalk or no-crosswalk), and 10 times higher learning rate when compared to the previous layers (due to fine-tuning).

\section{Experimental Methodology}
\label{sec:experiments}

In this section, we present the methodology used to evaluate the proposed system. First, the dataset is properly introduced and described. Then, the metrics used to evaluate the proposed system are presented. Finally, the experiments are detailed.

\subsection{Dataset}
\begin{figure}
	\centering
	\includegraphics[width=0.485\textwidth]{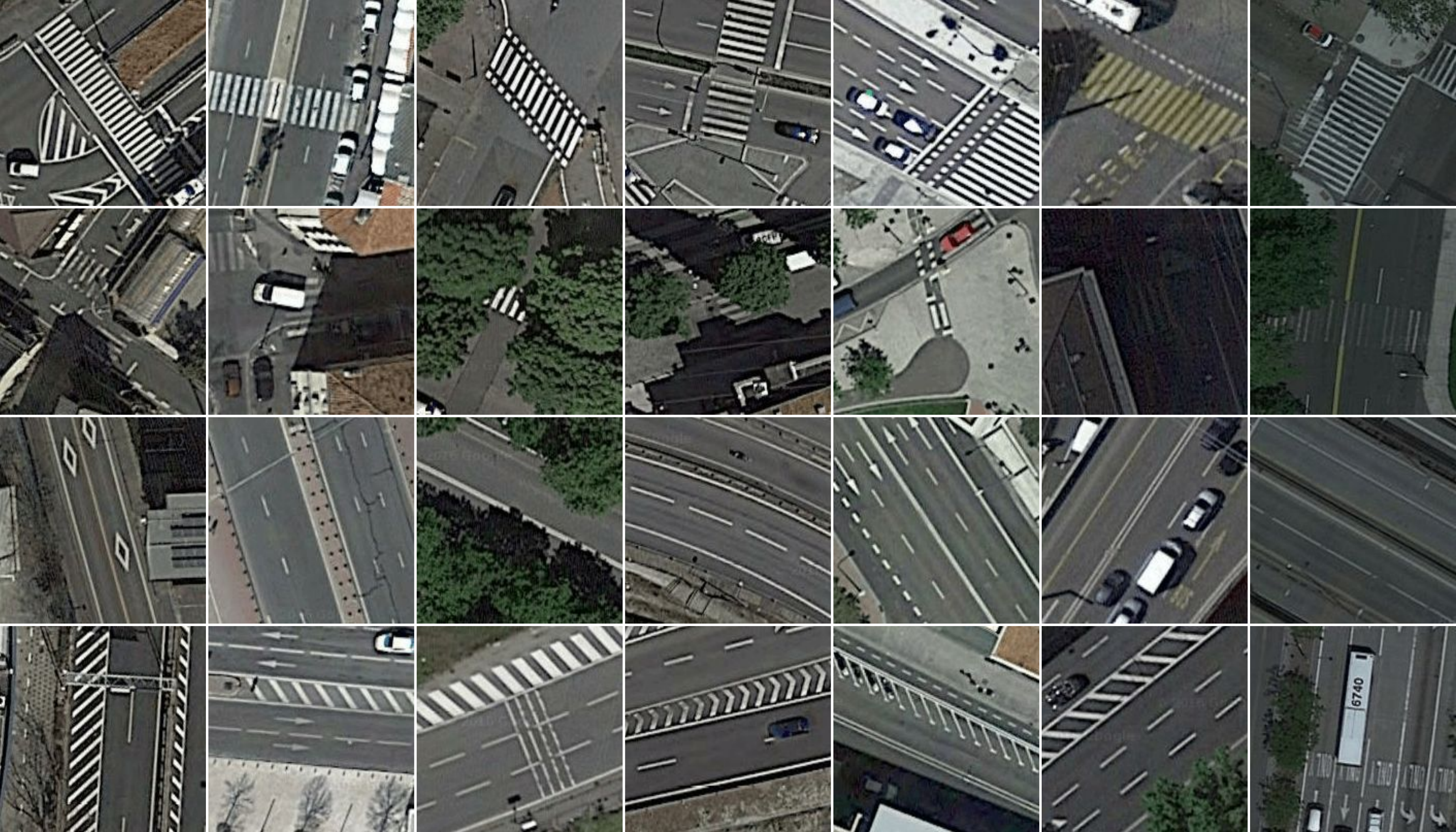}
	\caption{First row presents positive samples with varying layouts of crosswalks. Second row has some challenging positive cases (crosswalks with strong shadows, occlusions, aging, truncated, etc.). Third row presents different negative examples. Last row has some challenging negative cases.}
	\label{fig:samples}
\end{figure}

The dataset used in this work was automatically acquired and annotated using the system hereby presented. The system downloads satellite images using the \GoogleStaticMapsAPI and acquires the annotations using the \OpenStreetMap. In total, the dataset comprises 245,768 satellite images, 74,047 images of which contain crosswalks (positive samples) and 171,721 do not contain zebra crossings (negative samples). To the best of our knowledge, this is the largest satellite dataset for crosswalk-related tasks in the literature. In the wild, crosswalks can vary across different cities, different countries and different continents. Alongside the design variations, they can be presented in a variety of conditions (e.g. occluded by trees, cars, pedestrians; with painting fading away; with shadows; etc.). In order to capture all this diversity, this dataset comprises satellite imagery from 3 continents, 9 countries, and at least 20 cities. The cities were chosen considering the density of available annotations and the size of the city. It was given preference to big cities assuming that they are better annotated. In total, these images add up to approximately 135,000 square kilometers, even though different images may partially contain a shared area. Some samples of crosswalks are shown in the \autoref{fig:samples}. A summary of the dataset can be seen at the \autoref{tab:datasets}. It is worth noting that, even though each part of the dataset is named after a city, some of the selected regions were large enough to partially include neighboring towns. A more detailed description of each part of the dataset, region locations and scripts used for the data acquisition are publicly available\footnote{\url{http://github.com/rodrigoberriel/satellite-crosswalk-classification}}.

\begin{table}[!h]
	\centering
	\caption{Number of Images on the Datasets Grouped by Continents}
	\label{tab:datasets}
	\begin{tabular}{lccc}
		\toprule
		\textbf{Description}           & \textbf{Crosswalks} & \textbf{No-Crosswalks} & \textbf{Total} 	\\ \midrule
		Europe                		   & 42,554     		 & 99,461        		  & 142,015 		\\
		America               		   & 15,822     		 & 36,811        		  & 52,633 			\\
		Asia                  		   & 15,671    		 	 & 35,449        		  & 51,120  		\\ \midrule
		\textbf{Total}                 & \textbf{74,047}     & \textbf{171,721}       & \textbf{245,313}\\ \bottomrule
	\end{tabular}
\end{table}

\subsection{Metrics}
On this classification task, we reported the global accuracy (hits per number of images) and the $F_1$ score (harmonic mean of precision and recall).

\subsection{Experiments}

Several experiments were designed to evaluate the proposed system. Initially, three well-known Convolutional Neural Network architectures were evaluated: AlexNet, VGG and GoogLeNet. The images downloaded from the \GoogleStaticMapsAPI were upsampled from $200\times200$ to $256\times256$ using bilinear interpolation. As a data augmentation procedure, the images were randomly cropped ($224\times224$ to the VGG and GoogLeNet, and $227\times227$ to the AlexNet -- as in the original networks), and the images were randomly mirrored on-the-fly.

It is known that crosswalks vary across different cities, countries and continents. In this context, some experiments were designed based on the expectation that crosswalks belonging to the same locality (e.g., same city, same country, etc.) are more likely to present similar features. Therefore, in these experiments, some models were trained and evaluated on the same locality, and they were named with the prefix ``intra'': intra-city, intra-country, intra-continent and intra-world. At the smaller scales (e.g., city and country) only some of the datasets were used (assuming the performance may be generalized), at higher levels all available datasets were used. These datasets were randomly chosen considering all cities with high annotation density. Ultimately, the intra-world experiment was also performed.

Besides these intra-based experiments, in which samples tend to have high correlation (i.e., present more similarities), cross-based experiments were performed. The experiments named with the prefix ``cross'' are those in which the model was trained with a dataset and evaluated in another with the same level of locality, i.e., trained using data from a city and tested in another city. Cross-based experiments were performed on the three levels of locality: city, country and continent. Another experiment performed was the cross-level, i.e., the model trained using an upper level imagery (e.g., the world model) was evaluated in lower levels (e.g., continents).

All the experiments aforementioned share a common setup. Each dataset was divided into train, validation and test sets, with 70\%, 10\% and 20\%, respectively. For all the experiments, including the cross-based ones (cross-level included), none of the images in the test set were seen by the models during training or validation, i.e., train, validation and test sets were exclusive. In fact, the cross experiments used only the test-set of the respective dataset on which they were evaluated. This procedure enables fairer comparisons and conclusions about the robustness of the models, even though the entire datasets could have been used on the cross-based models. Regarding the training, all models were trained during 30 epochs and the learning rate was decreased three times by a factor of 10. The initial learning rate was set to $10^{-4}$ to all three networks.

Lastly, the annotations from the \OpenStreetMap may not be as accurate as required due to many factors (e.g., human error, zebra crossing was removed, etc.). Even though, we assume the vast majority of them are correct and accurate. To validate that, an experiment using manually labeled datasets from the three continents (America, Europe, and Asia -- 44,175 images in total) was performed. The experiment was designed to have three results: error of the automatic annotation; performance increase when switching from automatic labeled training data to manually labeled; and, correlation between the error of the automatic annotation and the error of the validation results.

\section{Results}

Several experiments were performed and their accuracy and $F_1$ score were reported. Initially, different architectures were evaluated on two levels of locality. As can be seen in the \autoref{tab:results:architectures}, VGG achieved the best results. It is interesting to notice that AlexNet, a smaller model that can be loaded into smaller GPUs, also achieved competitive results.

\begin{table}[h]
	\centering
	\caption{Different Architectures using Intra-Based Protocol}
	\label{tab:results:architectures}
	\begin{tabular}{ccccc}
		\toprule
		\textbf{Architecture}      & \textbf{Level} & \textbf{Dataset} & \textbf{Accuracy} & \textbf{$\bm{F_1}$ score} 	\\ \midrule
		\multirow{3}{*}{AlexNet}   & City    & Milan            & 96.69\%           & 94.56\%           	\\
		& City    & Turim            & 95.39\%           & 92.51\%           	\\
		& Country & Italy            & 96.06\%           & 93.53\%           	\\ \midrule
		\multirow{3}{*}{VGG}       & City    & Milan            & \textbf{97.00\%}  & \textbf{95.10\%}  	\\
		& City    & Turim            & \textbf{96.37\%}  & \textbf{94.15\%}  	\\
		& Country & Italy            & \textbf{96.70\%}  & \textbf{94.66\%}  	\\ \midrule
		\multirow{3}{*}{GoogLeNet} & City    & Milan            & 96.04\%           & 93.41\%           	\\
		& City    & Turim            & 94.27\%           & 90.78\%           	\\
		& Country & Italy            & 95.22\%           & 92.17\%           	\\ \bottomrule
	\end{tabular}
\end{table}

Regarding the intra-based experiments, the chosen model showed to be very consistent across the different levels of locality. The model achieved 96.9\% of accuracy (on average) and the details of the results can be seen in the \autoref{tab:results:intra}.

\begin{table}[h]
	\centering
	\caption{Intra-Based Results for the VGG Network}
	\label{tab:results:intra}
	\begin{tabular}{cccc}
		\toprule
		\textbf{Level}             & \textbf{Dataset} & \textbf{Accuracy} & \textbf{$\bm{F_1}$ score} \\ \midrule
		\multirow{2}{*}{City}      & Milan            & 97.00\%           & 95.10\%           \\
		& Turim            & 96.37\%           & 94.15\%           \\ \midrule
		\multirow{2}{*}{Country}   & Italy            & 96.70\%           & 94.66\%           \\
		& France           & 95.87\%           & 93.12\%           \\ \midrule
		\multirow{3}{*}{Continent} & Europe           & 96.72\%           & 94.50\%           \\
		& America          & 96.77\%           & 94.55\%           \\
		& Asia             & 98.61\%           & 97.71\%           \\ \midrule
		World                      & World            & 97.11\%           & 95.17\%           \\ \bottomrule
	\end{tabular}
\end{table}

\begin{table}[b]
	\centering
	\caption{Cross-Based Results for the VGG Network}
	\label{tab:results:cross-based}
	\begin{tabular}{ccccc}
		\toprule
		\textbf{Level}             & \textbf{Train/Val} & \textbf{Test} & \textbf{Accuracy} & \textbf{$\bm{F_1}$ score} \\ \midrule
		\multirow{2}{*}{City}      & Milan              & Turim         & 93.47\%           & 89.80\%           \\
		& Turim              & Milan         & 95.40\%           & 92.36\%           \\ \midrule
		\multirow{2}{*}{Country}   & Italy              & France        & 94.78\%           & 91.16\%           \\
		& France             & Italy         & 94.78\%           & 91.61\%           \\ \midrule
		\multirow{2}{*}{Continent} & Europe             & Asia          & 96.62\%           & 94.33\%           \\
		& Asia               & Europe        & 93.65\%           & 88.94\%           \\ \bottomrule
	\end{tabular}
\end{table}

As expected, the cross-based models achieved a lower overall accuracy when compared to the intra-based. Nevertheless, these models were still able to achieve high accuracies ($94.8\%$ on average, see \autoref{tab:results:cross-based}). This can be partially explained by inherent differences on the images between places far apart, i.e., different solar elevation angles cause notable differences on the images; the quality of the images may differ between cities; among other factors that tend to be captured by the model during the training phase.

Cross-level experiments reported excellent results. As already discussed, none of these models had contaminated test sets. Yet, on average, they achieved 96.1\% of accuracy. The robustness of these models can be seen in the \autoref{tab:results:cross-level}, where all the cross-level results were summarized.

\begin{table}[h]
	\centering
	\caption{Cross-Level Results for the VGG Network \protect\linebreak Train/Val$\rightarrow$Test}
	\label{tab:results:cross-level}
	\begin{tabular}{ccccc}
		\toprule
		\textbf{Cross-Level}                           & \textbf{Train/Val} & \textbf{Test} & \textbf{Accuracy} & \textbf{$\bm{F_1}$ score} \\ \midrule
		\multirow{2}{*}{Country$\rightarrow$City}      & Italy              & Milan         & 97.17\%           & 95.37\%           \\
		& Italy              & Turim         & 96.28\%           & 94.04\%           \\ \midrule
		\multirow{2}{*}{Continent$\rightarrow$Country} & Asia               & Portugal      & 92.71\%           & 86.04\%           \\
		& Asia               & Italy         & 94.69\%           & 91.43\%           \\ \midrule
		\multirow{3}{*}{World$\rightarrow$Continent}   & World              & Europe        & 96.72\%           & 94.50\%           \\
		& World              & America       & 96.66\%           & 94.35\%           \\
		& World              & Asia          & 98.65\%           & 97.79\%           \\ \bottomrule
	\end{tabular}
\end{table}

\begin{figure}[t]
	\centering
	\includegraphics[width=0.485\textwidth,height=0.07\textheight]{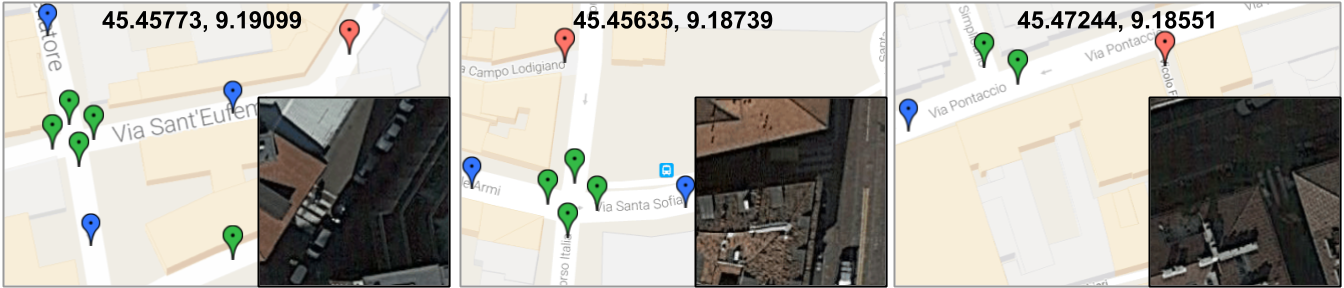}
	\caption{Failure cases. True Positive and True Negative are represented by green and blue markers, respectively. False Positive and False Negative are both represented by the red markers.}
	\label{fig:errors}
\end{figure}

Lastly, results of manual annotations showed the proposed system can automatically acquire and annotate satellite imagery with an average accuracy of $95.41\%$ ($4.04\%$ false positive and $4.83\%$ false negative samples), see \autoref{tab:results:human-annotation} for detailed automatic annotation errors. In addition, results of the manual annotation also showed a small improvement (2.00\% on average, see \autoref{tab:results:human-annotation}) on the accuracy when switching from automatic labeled training data to manually labeled. This improvement is due to the decrease in noise of models trained using the manual labels. Some failure cases are shown in \autoref{fig:errors}. As can be seen in the \autoref{tab:results:human-annotation}, there is a correlation between the error of the automatic annotation and the accuracy of the resulting models, i.e., an increase in the annotation error implies in a decrease in the accuracy of models using automatic data. The \autoref{tab:results:human-annotation} also shows that the absolute differences between models validated with automatic data and manual data are not very high, at most 1.75\% which is the difference for New York. This indicates that all our previous results of experiments evaluated with automatic data are valid, and would not be much different if they were evaluated with manually annotated data.

\begin{table}[h]
	\centering
	\caption{Impact of Manual Annotation on the Accuracy for the VGG \protect\linebreak A: Automatic -- M: Manual}
	\label{tab:results:human-annotation}
	\begin{tabularx}{0.45\textwidth}{*5{>{\centering\arraybackslash}X}} \toprule
		\multirow{2}{*}{\textbf{Dataset}} 
		& \multirow{2}{*}{\textbf{\begin{tabular}[c]{@{}c@{}}Annotation\\ Error\end{tabular}}} 
		& \multicolumn{3}{c}{\textbf{TrainVal / Test}} 
		\\ \cmidrule(l){3-5} 
		
		&  
		& \textbf{A / A}
		& \textbf{A / M}
		& \textbf{M / M}
		\\ \midrule
		Milan            & 2.58\%  & 97.00\%             & 97.71\%            & 98.91\%            \\
		Turim            & 6.57\%  & 96.37\%             & 94.69\%            & 98.32\%            \\
		New York         & 6.77\%  & 95.49\%             & 93.74\%            & 96.62\%            \\
		Toyokawa         & 1.47\%  & 98.16\%             & 99.04\%            & 99.33\%            \\ \bottomrule
	\end{tabularx}
\end{table}

\section{Conclusion}

In this letter, a scheme for automatic large-scale satellite zebra crossing classification was proposed. The system automatically acquires images of crosswalks and no-crosswalks around the world using the \GoogleStaticMapsAPI, \GoogleDirectionsAPI and \OpenStreetMap. Additionally, deep-learning-based models are trained and evaluated using these automatically annotated images. Experiments were performed on this novel dataset with 245,768 images from 3 different continents, 9 countries and more than 20 cities. Experimental results validated the robustness of the proposed system and showed an accuracy of 97.11\% on the global experiment.

\section*{Acknowledgment}

We would like to thank UFES for the support, CAPES for the scholarships, and CNPq (311120/2016-4) for the grant. We gratefully acknowledge the support of NVIDIA Corporation with the donation of the Tesla K40 GPU used for this research.

\bibliographystyle{IEEEtran}
\bibliography{GRSL2017}

\end{document}